\documentclass[conference]{IEEEtran}
\IEEEoverridecommandlockouts

\usepackage{cite} 
\usepackage[none]{hyphenat}

\usepackage{amsmath,amssymb,amsfonts}
\usepackage{algorithmic} % From IEEE template
\usepackage{graphicx}    % For \includegraphics
\usepackage{textcomp}    % From IEEE template
\usepackage{xcolor}
\usepackage{placeins}    % For \FloatBarrier
\usepackage{pgfplots}    % For plotting
\pgfplotsset{compat=1.18}
\usepackage{siunitx}
\usepackage{booktabs}
\usepackage{adjustbox}   % For scaling tables
\usepackage{stfloats}
\usepackage{dsfont}
\usepackage[T1]{fontenc} % Essential for vector fonts
\def\BibTeX{{\rm B\kern-.05em{\sc i\kern-.025em b}\kern-.08em
    T\kern-.1667em\lower.7ex\hbox{E}\kern-.125emX}}

\newcommand{\minkset}{|\mathcal{T}_{\min\text{-}k\%}|} % Use \mathcal for sets
\newcommand{\suminkset}{\sum_{x_t \in \mathcal{T}_{\min\text{-}k\%}}}
\newcommand{\meanlogprob}{\mu_{p, x_{<t}}} % Clarify context for mean/std
\newcommand{\stdlogprob}{\sigma_{p, x_{<t}}}  % Clarify context for mean/std
\newcommand{\Mp}{M_P}
\newcommand{\Mh}{M_H}
\newcommand{\LLcond}[2]{\text{LL}(#1 | #2)}
\newcommand{\LLuncond}[1]{\text{LL}(#1)}

\usepackage[hidelinks]{hyperref}         % hyperlinks
\usepackage{url}              % simple URL typesetting

\newtheorem{definition}{Definition}

\begin{document}

\title{On the Effectiveness of Membership Inference in Targeted Data Extraction from Large Language Models}

\author{
    \IEEEauthorblockN{Ali Al Sahili}
    \IEEEauthorblockA{\textit{American University of Beirut}\\
    Beirut, Lebanon \\
    ama318@mail.aub.edu}
    \and
    \IEEEauthorblockN{Ali Chehab}
    \IEEEauthorblockA{\textit{American University of Beirut}\\
    Beirut, Lebanon \\
    Chehab@aub.edu.lb}
    \and
    \IEEEauthorblockN{Razane Tajeddine}
    \IEEEauthorblockA{\textit{American University of Beirut}\\
    Beirut, Lebanon \\
    Razane.Tajeddine@aub.edu.lb}
}

\maketitle
\thispagestyle{plain} % For the first page
\pagestyle{plain}    % For subsequent pages

\begin{abstract}
Large Language Models (LLMs) are prone to memorizing training data, which poses serious privacy risks. Two of the most prominent concerns are training data extraction and Membership Inference Attacks (MIAs). Prior research has shown that these threats are interconnected: adversaries can extract training data from an LLM by querying the model to generate a large volume of text and subsequently applying MIAs to verify whether a particular data point was included in the training set. In this study, we integrate multiple MIA techniques into the data extraction pipeline to systematically benchmark their effectiveness. We then compare their performance in this integrated setting against results from conventional MIA benchmarks, allowing us to evaluate their practical utility in real-world extraction scenarios.
\end{abstract}


\section{Introduction}
Large Language Models (LLMs) have demonstrated remarkable proficiency in tackling a wide range of natural language processing (NLP) tasks \cite{achiam2023gpt4, anil2024gemini, dubey2024llama} and generalizing beyond language-related tasks. These models extend beyond text generation to support diverse and complex applications, including reasoning agents \cite{gao2024agentbased, wang2024autonomous}, code synthesis \cite{fan2024survey}, and multi-modal understanding \cite{wang2024survey, yin2024survey}.

Despite their capabilities, machine learning models in general, and specifically LLMs, are known to potentially expose information about their training data. Various vulnerabilities and privacy concerns have been identified in machine learning models \cite{song2020systematic}. This includes the risk of exposure to privacy attacks, such as membership inference \cite{shokri2017membership, hu2021survey} and data extraction \cite{fredrikson2015model}.

The association of privacy leakage with model overfitting \cite{xie2020artificial, carlini2019secret} has led to the initial assumption that LLMs are unlikely to leak sensitive data, given their strong generalization and large-scale training paradigm \cite{radford2019language, brown2020language, gaopile}. However, prior research has shown that LLM training data extraction is indeed feasible. A seminal study by Carlini et al. \cite{carlini2021extracting} proposed a method in which the model is prompted to produce large volumes of text, followed by the application of Membership Inference Attacks (MIAs) to identify candidate training sequences. Building on this foundation, more recent work has advanced MIA methodologies, introducing refined techniques and novel approaches tailored to enhance their effectiveness against LLMs \cite{mattern2023membership, shi2023detecting, fu2024membership}.

\begin{figure}[t]
\centering
\includegraphics[width=\columnwidth]{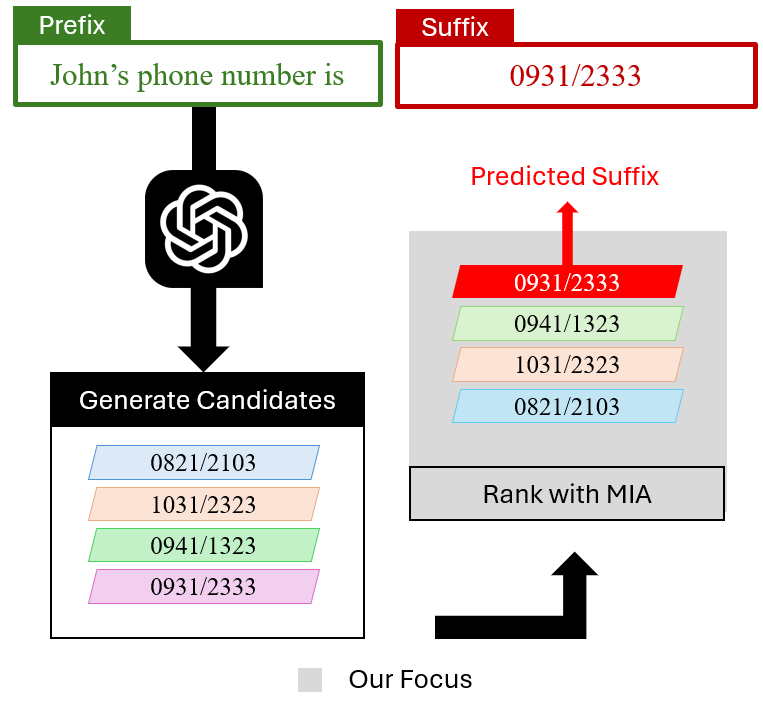}
\caption{An overview of the targeted data extraction pipeline, which is the focus of our investigation. An adversary provides a prefix, generates candidate suffixes, and then uses Membership Inference Attack (MIA) to rank them and predict the most likely continuation.}
\label{fig:overview}
\end{figure}

In this work, we examine the integration of various MIA techniques into the data extraction pipeline. Our evaluation focuses on how these methods affect the effectiveness of identifying verbatim training data extracted from LLMs.

\section{Related Work}
\subsection{Membership Inference Attacks and Data Extraction in ML}
Membership Inference Attacks (MIAs) aim to determine whether a particular data record was included in the training set of a model. The seminal work of Shokri et al. \cite{shokri2017membership} introduced black-box MIAs, where an adversary relies solely on query access to the target model. Their approach employed shadow models---trained on datasets with known membership---to capture distinguishing patterns between training and non-training samples. Building on this, Carlini et al. \cite{carlini2022membership} proposed the Likelihood Ratio Attack (LiRa), which enhances inference accuracy by comparing the likelihood of a sample under the training distribution against that of the non-training distribution. By leveraging the log-likelihood ratio, LiRA achieves strong attack performance while maintaining a low false-positive rate.

Data extraction attacks target ML models to recover training data by analyzing their outputs. Fredrikson et al. \cite{fredrikson2015model} showed that an attacker could reconstruct sensitive patient images from the predictions of a trained model in a medical diagnosis setting, demonstrating the risk of inversion attacks. Similarly, Song et al. \cite{song2017machine} introduced a method where models trained on sensitive data could be manipulated to output memorized training examples by carefully crafting input queries. These studies highlight the broader privacy vulnerabilities in ML beyond language models.

\subsection{Membership Inference Attacks Specifically for LLMs}
Prior research has examined MIAs in the context of LLMs. Carlini et al. \cite{carlini2021extracting} evaluated several metrics for conducting MIAs on data generated by GPT-2. Building on this, Mattern et al. \cite{mattern2023membership} proposed neighborhood attacks, a reference-free approach that assesses membership by comparing model scores of a sample against those of synthetically generated neighboring texts. Shi et al. \cite{shi2023detecting} introduced MIN-K\% PROB, which determines membership by analyzing the \(k\%\) of tokens with the lowest likelihoods, and further contributed a benchmark based on Wikipedia, where member and non-member sets were defined by article publication dates relative to model training. More recently, Xie et al. \cite{xie2024recall} presented ReCaLL, which infers membership by conditioning text on non-member prefixes and detecting shifts in likelihood.

In contrast, Duan et al. \cite{duan2024membership} introduced a benchmark that defines members as samples drawn from The Pile training set \cite{gaopile} and non-members from its test set. Their evaluation suggested that existing MIA techniques are generally weak, often performing close to random guessing. However, subsequent critiques argued that this setup diverges from the standard membership inference paradigm, as it deliberately minimizes distributional differences between members and non-members. Complementing this perspective, Meeus et al. \cite{meeus2025sok} showed that widely used post-hoc benchmarks suffer from substantial distribution shifts, enabling even a simple bag-of-words classifier to achieve near-perfect accuracy without relying on the model itself. They contend that such flaws undermine many reported successes of MIAs, attributing them to the exploitation of temporal or dataset artifacts rather than genuine memorization.

Offering a more nuanced perspective, Chen et al. \cite{chen2024statistical} provided a statistical and multi-perspective revisiting of MIAs, finding that while overall performance is often low, a significant number of "differentiable outliers" exist,  where member and non-member instances are easily distinguishable. Their work suggests that the inconsistency in prior research stems from the high dependency of MIA success on specific contexts like data domain and model size, rather than a universal failure of the attacks. Hayes et al. \cite{hayes2025strong} scaled one of the strongest MIAs (LiRA) to models up to 1B parameters. Their findings confirmed that strong attacks can succeed, but their effectiveness is limited in practical, single-epoch training scenarios. They also demonstrated that a sample's vulnerability to membership inference does not directly correlate with its risk of extraction, suggesting these two privacy threats capture different aspects of memorization.

\subsection{Data Extraction from LLMs}
Several methods exist for extracting verbatim training data from LLMs. Carlini et al. \cite{carlini2021extracting} introduced a technique that prompts the model to generate extensive text and then using MIA methods to filter for potential training data sequences. Their method involved text generation strategies like top-$n$ sampling and conditioning on Internet text prefixes, followed by ranking generated samples using metrics like perplexity ratios. This work demonstrated the extraction of sensitive information, highlighting the vulnerability of even large models. Carlini et al. \cite{carlini2022quantifying} quantified memorization in language models, demonstrating that memorization increases with 1) model capacity, 2) the repetition of training examples, and 3) the amount of context provided. They showed that LLMs memorize a significant portion of their training data. Nasr et al. \cite{nasr2023scalable} demonstrated the extraction of gigabytes of training data from various language models and developed a divergence attack to circumvent alignment in models like ChatGPT, increasing the extraction rate of training data. Most related to our work, Yu et al. \cite{yu2023bag} explored tricks for text generation and text ranking in data extraction. 

In this work, we make the following contributions:
\begin{itemize}
    \item We extensively study the effectiveness of membership inference methods in the targeted data extraction pipeline.
    \item We explore the possibility of reducing false extraction rates using these methods.
\end{itemize}

\section{Preliminaries}
\subsection{Defining Memorization}
Following the definition by Carlini et al. \cite{carlini2021extracting}, we use $k$-eidetic memorization, which is defined as follows:
\begin{definition}[$k$-eidetic memorization \cite{carlini2021extracting}]
    A string $s$ is $k$-eidetic memorized (for $k \geq 1$) by an LM $f_\theta$ if $s$ is extractable from $f_\theta$ and appears in at most $k$ examples in the training data $X$:
\begin{equation}
\begin{split}
|\{x \in X : s \subset x\}| \leq k.
\end{split}
\end{equation}
\end{definition}

This definition considers the number of distinct training examples containing a string. Lower values of $k$ suggest more potentially problematic memorization. For instance, memorizing a frequently occurring word (high $k$) is less concerning than memorizing a unique name and phone number (low $k$) found only in a few training documents, thus raising more serious privacy concerns.

\subsection{Basic Setups}
\paragraph{Dataset}
The dataset used in this study is a subset of the one defined in the LM Extraction Challenge\cite{lmextractionchallenge2021}, which is, to our knowledge, the only recognized benchmark specifically tailored to the task of targeted data extraction from LLMs. It consists of sampled examples from The Pile training dataset \cite{gaopile}, each comprising a 50-token prefix and a 50-token suffix. The task is to predict the suffix given the prefix. All 100-token sequences in this evaluation subset appear only once in the original training set (1-eidetic). We use a subset of 1000 examples from the dataset extraction challenge in our experiments, unless otherwise specified. 

\paragraph{Language Model}
Following the setup of the Data Extraction Challenge, we employ the GPT-Neo-1.3B model \cite{black2021gpt}, An open-source transformer replicating GPT-3, trained by EleutherAI on The Pile. GPT-Neo is an autoregressive language model $f_\theta$ that generates a sequence of tokens $x_0, x_1, \ldots, x_N$ via the chain rule:
\begin{equation}
\begin{split}
p(x_0, x_1, \ldots, x_N | \theta) = \prod_{n=0}^{N} p(x_n | x_{<n}, \theta),
\end{split}
\end{equation}
where $x_{[0,n-1]} = x_{<n} = \{x_0, \dots, x_{n-1}\}$ and $p(x_n | x_{<n}, \theta)$ is the probability of token $x_n$ given the preceding tokens $x_{<n}$ and model parameters $\theta$. Given a prefix $p$, we denote the probability of generating a certain suffix $s$ conditional on the prefix $p$ as $f_{\theta}(s|p)$.
\subsection{Threat Model}
We consider an adversary with black-box access to the target LLM. In this setting, the adversary can query the model with arbitrary inputs (prompts or prefixes) and observe the generated text outputs along with the associated probabilities or log-likelihoods of the tokens. However, the adversary cannot access the model’s weights, architecture details beyond publicly available information, or the training dataset.

The primary goal of the adversary in the context of training data extraction is to recover verbatim sequences from the LLM's training data. Specifically, for targeted extraction, it is assumed that the adversary has a prefix $p_{true}$ that is known or suspected to be part of a training sequence. The objective is then to prompt the LLM with $p_{true}$ and, from the generated continuations (candidate suffixes), identify the true suffix $s_{true}$ such that the concatenated sequence $[p_{true}, s_{true}]$ was an exact segment present in the training data. This is more challenging than non-targeted extraction, where any memorized sequence is of interest. The success of the attack hinges on the model's tendency to memorize and the adversary's ability to effectively generate and rank candidate sequences to pinpoint these memorized instances.

\subsection{Evaluation Metrics}
To evaluate the performance of our data extraction pipeline, we adopt two metrics: precision ($M_P$), and Hamming distance ($M_H$). These metrics capture exact-match accuracy and token-level similarity.

\paragraph{Precision ($M_P$)}
We define $M_P$ as the proportion of correctly extracted suffixes among the top-$1$ ranked outputs for each prefix. A suffix is considered correctly extracted only if it exactly matches the ground-truth suffix.
\begin{equation}
\begin{split}
M_P = \frac{1}{N} \sum_{i=1}^{N} \mathds{1}[\hat{y}_i^{(1)} = y_i],
\end{split}
\end{equation}
where $N$ is the total number of prefixes, $\hat{y}_i^{(1)}$ is the generated top-$1$ suffix for the $i$-th prefix, and $y_i$ is the corresponding ground-truth suffix.

\paragraph{Hamming Distance $M_H$}
To assess partial similarity at the token level, we compute the average Hamming distance between the top-1 generated suffix and the ground truth. The Hamming distance is defined as the number of differing tokens at corresponding positions.

\begin{equation}
\begin{split}
M_H = \frac{1}{N} \sum_{i=1}^{N} \left( \frac{1}{L} \sum_{j=1}^{L} \mathds{1}[x_{i,j} \ne y_{i,j}] \right),
\end{split}
\end{equation}

where $x_{i,j}$ and $y_{i,j}$ denote the $j$-th tokens of the top-1 generated suffix ($x_i = \hat{y}_i^{(1)}$) and ground-truth suffix ($y_i$)  of length $L$, respectively.

\section{Data Extraction Pipeline}
\subsection{Pipeline Overview}
The standard pipeline for targeted data extraction is a two-stage process. First, an adversary prompts the model with known prefixes to generate a large set of candidate continuations (suffixes). Second, the adversary uses a scoring function---typically a Membership Inference Attack (MIA) method---to rank these candidates, with the goal of identifying the one most likely to be a verbatim training sequence. While our work focuses on suffix ranking and the effectiveness of MIAs in this step, we still employ several suffix generation strategies. The goal is to obtain a varied set of generated data while avoiding unintended bias when evaluating MIAs in the subsequent stage.  

\subsection{Generation Methods}
We employ several techniques to enhance suffix generation beyond simple greedy decoding or basic sampling. These methods were thoroughly studied by Yu et al. \cite{yu2023bag} in a similar experimental setup to our work, where the authors evaluated them on the data extraction benchmark with extensive hyperparameter tuning and variant settings.
\par\medskip
\noindent\textbf{Top-k Sampling ($k$) \cite{fan2018hierarchical}:} At each generation step, the model considers only the $k$ tokens with the highest probabilities for sampling. This prunes the long tail of low-probability tokens.
\par\medskip
\noindent\textbf{Nucleus Sampling ($\eta$) \cite{holtzman2019curious}:} Also known as top-p sampling, this method selects the smallest set of tokens whose cumulative probability mass exceeds $p_{nucleus}$. The next token is then sampled from this dynamic set. This adapts the vocabulary size based on the model's certainty.
\par\medskip
\noindent\textbf{Typical Sampling ($\phi$) \cite{meister2023locally}:} This strategy aims to select tokens whose information content (negative log-probability) is close to the expected information content of the distribution. It tries to avoid both very high and very low probability tokens if they are atypical for the current context.
\par\medskip
\noindent\textbf{Temperature ($T$) \cite{hinton2014distilling}:} Temperature scaling adjusts the sharpness of the probability distribution over the vocabulary. A temperature $T < 1$ makes the distribution sharper, increasing the likelihood of high-probability tokens, while $T > 1$ flattens the distribution, making sampling more random and diverse.
\par\medskip
\noindent\textbf{Repetition Penalty ($r$) \cite{keskar2019ctrl}:} This technique modifies the logits of tokens that have already appeared in the generated sequence (or context) to discourage or encourage repetition. Typically, a penalty $r > 1$ divides the logit of a repeated token, making it less likely to be sampled again.
\par\medskip
\subsection{Ranking Methods}
We evaluate various MIA techniques as ranking scores. For all methods, a higher score is designed to indicate a higher likelihood of membership. $p$ refers to the true prefix given for targeted extraction. $s$ is a candidate suffix.
\par\medskip
\noindent\textbf{Likelihood \cite{yeom2018privacy}:} This is the baseline ranking method, where suffixes with higher log-likelihood (lower perplexity) given $p$ are ranked higher. The score is 
\begin{equation}
\begin{split}
M_{\text{L}}(s|p) = \LLcond{s}{p}.
\end{split}
\end{equation}

\par\medskip
\noindent\textbf{Zlib Entropy (Zlib) \cite{carlini2021extracting}:} To avoid assigning high likelihood to trivial completions, Carlini et al. introduced this metric which normalizes the log-likelihood of the suffix $s$ (given $p$) by its Zlib-compressed length. A higher score is better, implying higher likelihood relative to complexity.
\begin{equation}
\begin{split}
 M_{\text{Zlib}}(s|p) = \frac{\LLcond{s}{p}}{\text{Zlib}(s)}.
\end{split}
\end{equation}
\par\medskip
\noindent\textbf{High Confidence\cite{yu2023bag}:} Yu et al. introduced this method to reward sequences generated with high certainty. This reward is achieved by adding a bonus to the score for each such high-confidence token.
\begin{equation}
\begin{split}
M_{\text{HC}}(s|p) = \frac{1}{|s|} \sum_{x_t \in s} \mathcal{L}_{\text{adj}}(x_t),
\end{split}
\end{equation}
\begin{equation}
\begin{split}
\mathcal{L}_{\text{adj}}(x_t) = \mathcal{L}(x_t) - (\mathbb{I}[\text{conf}_1] - \mathbb{I}[\text{conf}_2]) \cdot \alpha \cdot \bar{\mathcal{L}}_{\text{batch}}. \\
\end{split}
\end{equation}
\par\medskip
\noindent\textbf{Outlier-Robust Likelihood \cite{yu2023bag}:} To prevent outlier tokens from disproportionately influencing the score, Yu et al. introduced this approach which calculates Likelihood using only raw tokens with log-probabilities that fall within a specified range, and replaces remaining outliers with the mean of all tokens in the sequence.
\begin{gather}
\mathcal{L}'(x_t) =
\begin{cases} 
    \mathcal{L}(x_t) & \text{if } |\mathcal{L}(x_t) - \mu_{\mathcal{L}}| \leq 3\sigma_{\mathcal{L}} \\
    \mu_{\mathcal{L}} & \text{otherwise}
\end{cases} \\[6pt]
M_{\text{ORL}}(s|p) = \frac{1}{|s|} \sum_{x_t \in s} \mathcal{L}'(x_t).
\end{gather}
\par\medskip
\noindent\textbf{SURP (Surprising Tokens) \cite{zhang2024adaptive}:} Zhang et al. introduced this method, which identifies "surprising tokens" in $s$ and uses their average log-probability as the membership score. A token is considered surprising if the model is confident in its prediction, but assigns a low probability to the actual token. The score is higher for sequences where these surprising tokens are less surprising. Let $\mathcal{T}_{\text{SURP}}$ be the set of surprising tokens in $s$.
\begin{equation}
\begin{split}
 M_{\text{SURP}}(s|p) = \frac{1}{|\mathcal{T}_{\text{SURP}}|} \sum_{x_t \in \mathcal{T}_{\text{SURP}}} \log P(x_t | p, x_{<t}).
\end{split}
\end{equation}
\par\medskip

\noindent\textbf{ReCaLL (Relative Conditional Log-Likelihood) \cite{xie2024recall}:} Xie et al. found that conditioning the target on a generic non-member prefix causes a noticeable log-likelihood shift. This method examines the ratio of the conditional log-likelihood of the suffix $s$ when prefixed by a non-member context $p_{nm}$ to its unconditional log-likelihood.
\begin{equation}
\begin{split}
 M_{\text{R}}((p,s); p_{nm}) = \frac{\LLcond{(p,s)}{p_{nm}}}{\LLuncond{(p,s)}}.
\end{split}
\end{equation}
\par\medskip
\noindent\textbf{S-ReCaLL( Suffix ReCaLL ):} Built on the premise that a true prefix should be highly predictive of its memorized suffix, we introduce this method which calculates the ratio of the suffix's unconditional negative log-likelihood (NLL) to its conditional NLL given the true prefix $p$. This represents an extension of the ReCaLL method designed to exploit the true prefix as an additional source of information.
\begin{equation}
\begin{split}
 M_{\text{SR}}(s|p) = \frac{\LLcond{s}{p}}{\LLuncond{s}}.
\end{split}
\end{equation}
\par\medskip
\noindent\textbf{Con-ReCaLL \cite{wang2024con}:} Wang et al. expanded on the ReCaLL method by conditioning on generic members in addition to nonmembers. The score is based on the difference between the suffix's log-likelihood when conditioned on a non-member prefix ($p_{nm}$) and when conditioned on a member prefix ($p_{m}$), normalized by the unconditional log-likelihood.
\begin{equation}
\begin{split}
 M_{\text{CR}}(s|p; p_{nm}) = \frac{\gamma \cdot \LLcond{(p,s)}{p_{m}} - \LLcond{(p,s)}{p_{nm}}}{\LLuncond{(p,s)}}.
\end{split}
\end{equation}
\par\medskip

\noindent\textbf{Lowercase \cite{carlini2021extracting}}: Carlini et al. introduced this metric which measures the ratio of a model’s perplexity on an original suffix compared to its fully lowercased version. The intuition is that verbatim content often preserves distinctive capitalization patterns, and when these are normalized, the model is expected to assign them a substantially lower likelihood, thereby signaling memorization.
\par\medskip

\noindent\textbf{Min-k\% Prob (Min-K\%) \cite{shi2023detecting}:} Shi et al. based this method on the claim that even the least predictable parts of a memorized sequence should have high likelihood. The method calculates the average log probability of the $k\%$ tokens in $s$ (given $p$) with the lowest conditional probabilities. A higher score indicates stronger confidence even in the "hardest" parts of the sequence.
\begin{equation}
\begin{split}
 M_{\text{K\%}}(s|p) &= \frac{1}{\minkset} \\
 &\quad \times \suminkset \log P(x_t | p, x_{<t}).
\end{split}
\end{equation}
\par\medskip
\noindent\textbf{Min-K\%++ \cite{zhang2024min}:} Zhang et al. enhanced Min-K\% by normalizing token log probabilities using the mean $\meanlogprob$ and standard deviation $\stdlogprob$ of log probabilities over the vocabulary for the context $(p, x_{<t})$.
\begin{equation}
\begin{split}
 M_{\text{K\%+}}(s|p) &= \frac{1}{\minkset} \\
 &\quad \times \suminkset \left( \frac{\log P(x_t | p, x_{<t}) - \meanlogprob}{\stdlogprob} \right).
\end{split}
\end{equation}
\subsection{Results and Analysis}
We record the performance of various MIA methods in suffix ranking while varying different aspects of our experimental setup. The detailed hyperparameter configurations for all generation and ranking methods are provided in Appendix A.
\subsubsection{Impact of the Generation Strategy}
We start by varying the suffix generation strategy, which produces the candidate suffixes that are then ranked by MIA methods. For each generation technique, we use the hyperparameters that obtained the highest precision in the work of Yu et al. \cite{yu2023bag}, and generate 20 candidates for each prefix. We also use a multi-constraint sampling configuration which was auto-tuned by Yu et al. \cite{yu2023bag}. Table \ref{tab:gen_strategies_comparison} shows the results for each generation strategy when using the baseline ranking method (likelihood). We notice that the multi-constraint configuration corresponds to the highest ($\Mp$) and lowest ($M_{H}$), obtaining the highest number of recovered verbatim suffixes, which is consistent with the results obtained in previous work\cite{yu2023bag}. 

\begin{table}[!b]
\caption{Comparison of Suffix Generation Strategies on Extraction Metrics (Ranked by Likelihood)}
\label{tab:gen_strategies_comparison}
\centering
\begin{tabular}{l c  c}
\toprule
\textbf{Generation Strategy} & \textbf{$\Mp$(\%)} &  \textbf{$\Mh$} \\
\midrule
Nucleus Sampling (Top-p) & 49.6 & 16.353 \\
Temperature Sampling & 49.2 & 16.365 \\
Typical Sampling & 47.4 & 16.996 \\
Top-k Sampling & 39.7 & 19.725 \\
Repetition Penalty & 38.8 & 20.544 \\
Composite Improved Generation & \textbf{50.7} & \textbf{15.858}\\
\bottomrule
\end{tabular}
\end{table}
\begin{table*}[t]
\centering
\caption{Suffix Ranking Performance Across Different Generation Methods. Results were obtained after 5 trials}
\label{tab:mia_ranking_all_gens}
\footnotesize
\setlength{\tabcolsep}{2pt}
\begin{tabular}{@{}l@{\hspace{0.5em}}||S[table-format=2.1]|S[table-format=2.2] @{\hspace{0.5em}} S[table-format=2.1]|S[table-format=2.2] @{\hspace{0.5em}} S[table-format=2.1]|S[table-format=2.2] @{\hspace{0.5em}} S[table-format=2.1]|S[table-format=2.2] @{\hspace{0.5em}} S[table-format=2.1]|S[table-format=2.2] @{\hspace{0.5em}} S[table-format=2.1]|S[table-format=2.2]@{}}
\toprule
\textbf{MIA Ranker} & \multicolumn{2}{c}{\textbf{Nucleus (top-p)}} & \multicolumn{2}{c}{\textbf{Temperature}} & \multicolumn{2}{c}{\textbf{Typical}} & \multicolumn{2}{c}{\textbf{Top-k}} & \multicolumn{2}{c}{\textbf{Rep. Penalty}} & \multicolumn{2}{c}{\textbf{Composite}} \\
\cmidrule(lr){2-3} \cmidrule(lr){4-5} \cmidrule(lr){6-7} \cmidrule(lr){8-9} \cmidrule(lr){10-11} \cmidrule(lr){12-13}
& {$\Mp$(\%)} & {$\Mh$} & {$\Mp$(\%)} & {$\Mh$} & {$\Mp$(\%)} & {$\Mh$} & {$\Mp$(\%)} & {$\Mh$} & {$\Mp$(\%)} & {$\Mh$} & {$\Mp$(\%)} & {$\Mh$} \\
\midrule
Likelihood        & 50.5 & 15.91 & 50.8 & 15.81 & 50.2 & 15.95 & 46.5 & 17.32 & 45.0 & 17.84 & 50.8 & 15.75 \\
Zlib              & 49.8 & 16.17 & 50.4 & 15.85 & 49.6 & 16.16 & 46.4 & 17.38 & 45.0 & 17.85 & 50.1 & 16.01 \\
Outlier            & 49.1 & 16.24 & 49.2 & 16.11 & 48.2 & 16.66 & 43.6 & 18.24 & 42.3 & 18.61 & 49.0 & 16.14 \\
SURP              & 48.4 & 16.70 & 49.4 & 16.29 & 47.8 & 16.93 & 45.2 & 17.80 & 43.6 & 18.26 & 49.1 & 16.45 \\
High Confidence   & 50.2 & 16.03 & 50.4 & 15.85 & 49.9 & 16.04 & \textbf{46.5} & 17.29 & 45.0 & 17.84 & 50.4 & 15.92 \\
ReCaLL            & \textbf{50.7} & 15.78 & 50.6 & 15.68 & 50.3 & \textbf{15.77} & 46.3 & \textbf{17.20} & 44.8 & \textbf{17.77} & 50.8 & 15.64 \\
S-ReCaLL          & 50.5 & \textbf{15.77} & \textbf{51.0} & \textbf{15.50} & 50.2 & 15.87 & 46.3 & 17.35 & \textbf{45.0} & 17.89 & 50.9 & \textbf{15.49} \\
Lowercase         & 44.4 & 18.06 & 43.0 & 18.29 & 41.7 & 19.09 & 27.2 & 24.43 & 27.1 & 24.40 & 43.7 & 17.99 \\
CON-ReCaLL        & 50.0 & 15.89 & 50.2 & 15.76 & 49.9 & 15.94 & 45.6 & 17.64 & 44.0 & 18.32 & 50.4 & 15.72 \\
Min-K\%           & 50.6 & 15.88 & 50.9 & 15.73 & \textbf{50.3} & 15.86 & 46.5 & 17.27 & 45.1 & 17.84 & \textbf{51.0} & 15.67 \\
Min-K\%++         & 47.6 & 17.04 & 47.3 & 16.85 & 46.1 & 17.42 & 36.6 & 20.44 & 35.5 & 20.90 & 47.3 & 16.94 \\
\bottomrule
\end{tabular}
\end{table*}
As outlined earlier, the purpose of employing multiple suffix generation techniques is to create a diverse pool of candidate suffixes for the ranking stage. Each set of generated suffixes is subsequently evaluated using various MIA methods, with performance measured by the accuracy of identifying the ground-truth suffix. The results, summarized in Table \ref{tab:mia_ranking_all_gens} for 20 candidate suffixes per generation technique, reveal that while certain methods (e.g., S-ReCaLL, Min K\%) achieve consistent but marginal gains over the baseline ranking, most approaches perform comparably to the baseline. In contrast, methods such as lowercase and Min-K\%++ systematically underperform. These findings indicate that the choice of MIA ranker offers limited potential for enhancing the overall effectiveness of the data extraction attack.

\subsubsection{Analysis of Generation Count}
\begin{figure}[b]
\centering
\includegraphics[width=0.7\columnwidth]{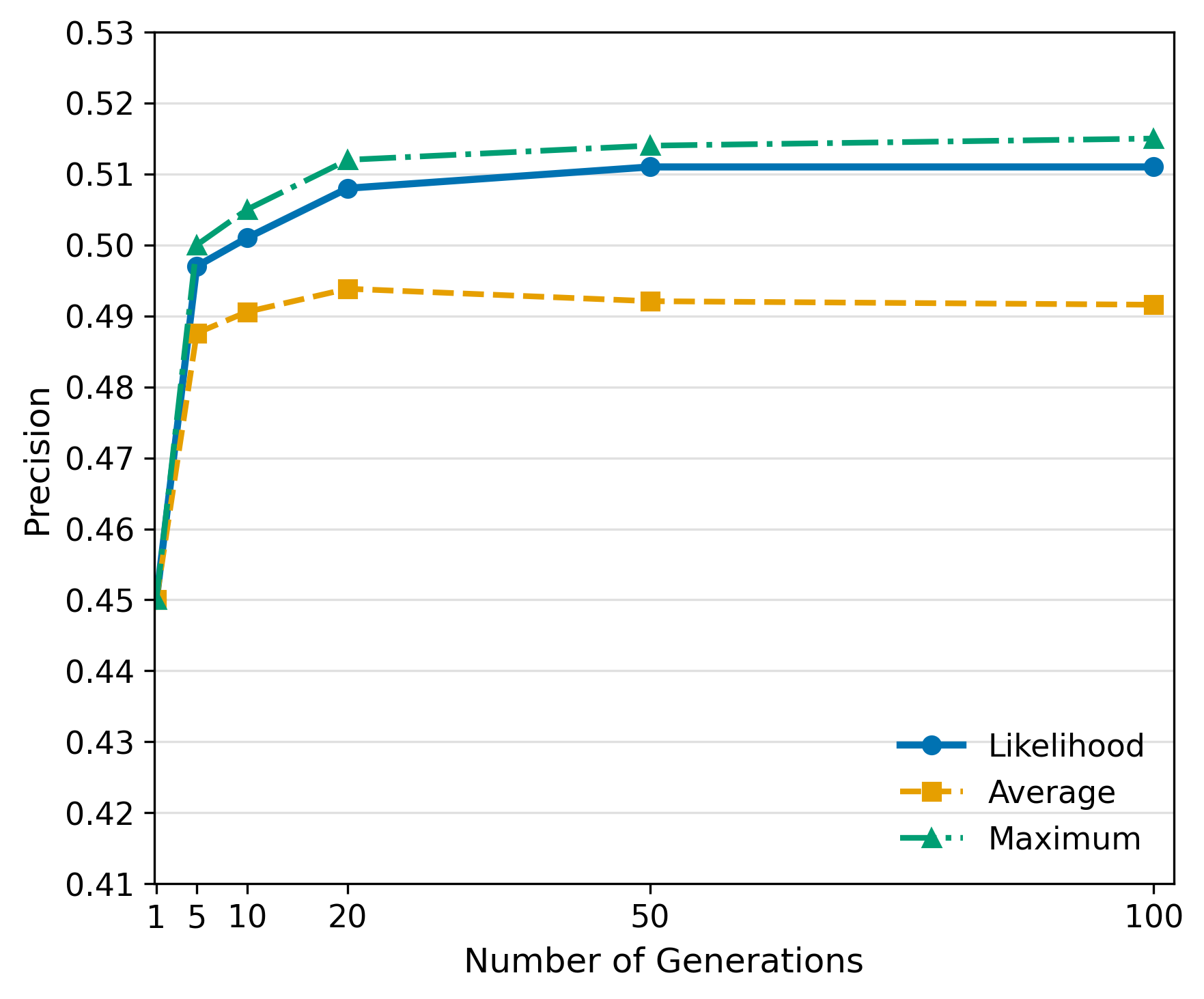}
\caption{Precision ($\Mp$) vs. number of generated candidates for various MIA rankers using the Composite Generation strategy. Note the diminishing returns as the number of candidates exceeds 50.}
\label{fig:precision_vs_generations}
\end{figure}

We next examine how the size of the candidate suffix pool influences the effectiveness of ranking techniques. Using the multi-constraint generation strategy, we evaluate each ranking method while varying the number of generated suffixes per prefix. As shown in Figure \ref{fig:precision_vs_generations}, the baseline precision, along with the mean and maximum precision of other methods, generally increases as the candidate pool expands, highlighting the benefit of generating a larger set of options: a broader pool raises the likelihood that the true suffix is available for ranking. However, these gains taper off beyond approximately 20 candidates, after which the mean performance stabilizes. Crucially, the central trend remains unchanged, as the best-performing ranking methods yield only marginal improvements over the baseline, regardless of pool size. A detailed breakdown of the precision and Hamming distance for each ranker across all candidate pool sizes is presented in Appendix B.1.

\subsubsection{Analysis of Target Model Architecture and Scale}
Although our primary experiments rely on the GPT-Neo 1.3B model to remain consistent with the data extraction challenge setup, model scale 
and architecture both play a critical role in the susceptibility to data extraction attacks. Larger models, with increased capacity, have been shown to exhibit greater vulnerability to memorization and verbatim leakage \cite{carlini2022quantifying},
 and different model architectures may lead to different memorization patterns and behavior. To assess the impact of model size on suffix ranking performance, we extend our evaluation to several models from the GPT-Neo family and related GPT architectures, spanning sizes from 125M to 6B parameters. To assess the impact of model architecture, we evaluate the attack on several models from the Pythia family, a suite of models also trained using The Pile dataset. For both experiments, we adopt the multi-constraint generation strategy and generate 20 candidate suffixes per prefix.

As shown in Table \ref{tab:target_model_effect}, extraction precision ($\Mp$\%) rises steadily with model size, providing strong evidence that larger models are more prone to reproducing training data verbatim.

Equally important, the results confirm earlier findings regarding the limited benefit of MIA ranking methods: across all model scales, the performance gains of the best-performing ranker over the simple likelihood baseline remain consistently marginal. This is also observed when applying the pipelines on Pythia models. For a comprehensive view, the full performance results for all evaluated MIA rankers across each model are available in Appendix B.2.

\begin{table}[t]
\caption{Analysis of Target Model Scale on Extraction Precision ($\Mp$\%).}
\label{tab:target_model_effect}
\centering
\begin{tabular}{l c c}
\toprule
\textbf{Model Size} & \textbf{Baseline Ranking ($\Mp$\%)} & \textbf{Highest Score ($\Mp$\%)} \\
\midrule
GPT-Neo 125M & 19.8 & 20.2 \\
GPT-Neo 1.3B & 50.8 & 51.3 \\
GPT-Neo 2.7B & 58.7 & 58.8 \\
GPT-J 6B   & \textbf{70.6} & \textbf{70.8} \\
\midrule
Pythia 410M & 29.8 & 30.2 \\
Pythia 1.4B & 48.9 & 49.2 \\
Pythia 2.8B & 56.0 & 56.3 \\
Pythia 6.9B & \textbf{62.5} & \textbf{62.5} \\
\bottomrule
\end{tabular}
\end{table}

\subsubsection{Performance on an Extended Subset}
To validate our findings on a larger scale, we replicated the extraction pipeline experiment on an extended subset of the extraction challenge dataset, consisting of 15,000 prefix-suffix pairs. For this experiment, we used the multi-constraint generation method to produce 20 candidate suffixes for each prefix. The performance of various MIA methods is shown in Table \ref{tab:extended_extraction_results}. 

As shown in the table, the relative performance of the MIA methods remains stable. S-ReCaLL achieves the highest precision and lowest Hamming distance, but its improvement over the simple Likelihood baseline is marginal, at approximately 0.2 percentage points. Most methods cluster closely around the baseline's performance, while some methods, like Lowercase and Min-K\%++, significantly underperform. This reinforces our earlier conclusion that while some MIA methods can offer a slight edge in ranking, their impact is limited, and a simple likelihood score remains a strong baseline for this task.

\begin{table}[t]
\centering
\caption{Suffix Ranking Performance on the Extended Dataset. }
\label{tab:extended_extraction_results}
\begin{adjustbox}{max width=\columnwidth}
\begin{tabular}{l c c}
\toprule
\textbf{MIA Ranker} & \textbf{Precision $\Mp$ (\%)} & \textbf{Hamming Distance $\Mh$} \\
\midrule
Likelihood        & 49.0 & 16.84 \\
Zlib              & 48.7 & 17.01 \\
Outlier            & 47.8 & 17.11 \\
High Confidence   & 48.7 & 16.95 \\
S-ReCaLL          & \textbf{49.2} & \textbf{16.75} \\
ReCaLL            & 48.6 & 16.90 \\
Lowercase         & 42.1 & 19.40 \\
CON-ReCaLL        & 48.1 & 17.09 \\
Min-K\%           & 49.0 & 16.80 \\
Min-K\%++         & 45.7 & 17.89 \\
SURP              & 48.3 & 17.10 \\
\bottomrule
\end{tabular}
\end{adjustbox}
\end{table}

\begin{table*}[t]
\centering
\caption{MIA Performance for Extraction Confirmation Across Generation Methods. Results were obtained after 5 trials.}
\label{tab:mia_confirmation_all_gens}
\adjustbox{max width=\textwidth}{
\begin{tabular}{@{}l|ccc|ccc|ccc@{}}
\toprule
& \multicolumn{3}{c|}{\textbf{Nucleus (top-p)}} & \multicolumn{3}{c|}{\textbf{Temperature}} & \multicolumn{3}{c}{\textbf{Typical}} \\
\cmidrule(lr){2-4} \cmidrule(lr){5-7} \cmidrule(lr){8-10}
\textbf{MIA Method} & AUROC & TPR@5 & FPR@95 & AUROC & TPR@5 & FPR@95 & AUROC & TPR@5 & FPR@95 \\
\midrule
Likelihood        & 82.9 & 37.5 & 51.3 & 82.7 & 37.2 & 52.6 & 83.3 & 37.6 & 49.2 \\
Zlib              & 82.5 & 36.9 & 52.4 & 82.3 & 36.6 & 53.4 & 83.0 & 37.5 & 50.6 \\
Outlier            & 80.0 & 25.8 & 53.7 & 79.8 & 25.6 & 54.5 & 80.5 & 27.2 & 52.8 \\
High Confidence   & 83.1 & 37.1 & 51.2 & 82.9 & 36.8 & 52.2 & 83.6 & 37.3 & 48.8 \\
S-ReCaLL          & \textbf{88.1} & \textbf{41.2} & \textbf{40.9} & \textbf{87.9} & \textbf{41.0} & \textbf{43.0} & \textbf{88.4} & \textbf{41.5} & \textbf{39.5} \\
ReCaLL            & 68.9 & 16.6 & 82.2 & 68.8 & 16.6 & 82.8 & 70.0 & 16.8 & 79.6 \\
Lowercase         & 69.2 & 16.3 & 81.9 & 68.7 & 17.1 & 84.1 & 68.2 & 16.7 & 84.1 \\
Con-ReCaLL        & 67.7 & 16.1 & 82.4 & 67.5 & 15.9 & 82.5 & 68.9 & 16.2 & 79.6 \\
Min-K\%           & 84.1 & 38.6 & 48.2 & 83.9 & 38.3 & 49.1 & 84.6 & 38.8 & 46.3 \\
Min-K\%++         & 56.4 & 6.7  & 82.2 & 56.2 & 6.7  & 82.2 & 57.5 & 6.8  & 79.3 \\
SURP              & 81.8 & 37.7 & 57.6 & 81.6 & 37.5 & 57.8 & 82.4 & 38.2 & 56.2 \\
\midrule
\midrule
& \multicolumn{3}{c|}{\textbf{Top-k}} & \multicolumn{3}{c|}{\textbf{Rep. Penalty}} & \multicolumn{3}{c}{\textbf{Composite}} \\
\cmidrule(lr){2-4} \cmidrule(lr){5-7} \cmidrule(lr){8-10}
\textbf{MIA Method} & AUROC & TPR@5 & FPR@95 & AUROC & TPR@5 & FPR@95 & AUROC & TPR@5 & FPR@95 \\
\midrule
Likelihood        & 87.0 & 41.9 & 40.4 & 88.4 & 43.4 & 37.5 & 82.6 & 37.2 & 52.3 \\
Zlib              & 86.8 & 41.7 & 41.1 & 88.3 & 44.3 & 38.1 & 82.2 & 36.6 & 53.2 \\
Outlier            & 84.2 & 34.9 & 45.9 & 85.6 & 37.6 & 42.0 & 79.6 & 25.6 & 54.9 \\
High Confidence   & 87.2 & 41.5 & 39.9 & 88.5 & 43.0 & 36.7 & 82.8 & 36.8 & 52.4 \\
S-ReCaLL          & \textbf{90.1} & \textbf{45.6} & \textbf{34.3} & \textbf{91.0} & \textbf{47.4} & \textbf{31.8} & \textbf{87.9} & \textbf{40.9} & \textbf{42.0} \\
ReCaLL            & 73.2 & 19.0 & 70.3 & 74.9 & 19.8 & 66.8 & 68.8 & 16.5 & 82.5 \\
Lowercase         & 69.8 & 17.5 & 82.9 & 70.7 & 17.9 & 81.8 & 68.7 & 17.0 & 83.8 \\
Con-ReCaLL        & 72.0 & 18.1 & 74.7 & 73.5 & 18.6 & 70.6 & 67.6 & 15.9 & 82.3 \\
Min-K\%           & 87.8 & 43.4 & 37.7 & 89.1 & 45.2 & 35.1 & 83.9 & 38.3 & 49.9 \\
Min-K\%++         & 63.6 & 8.5  & 67.0 & 66.2 & 9.2  & 62.2 & 55.7 & 6.5  & 84.1 \\
SURP              & 86.6 & 42.1 & 43.1 & 88.0 & 43.6 & 39.7 & 81.7 & 37.4 & 57.1 \\
\bottomrule
\end{tabular}
}
\end{table*}
\section{Mitigating False Extractions }
A key observation from our analysis of the two-stage pipeline is that even with the best combination of generation and ranking methods, the precision ($\Mp$) peaks at 51\%. This means that even under optimal conditions, nearly half of the top-ranked suffixes that are predicted as verbatim extraction are false positives. Although recovering more than half of the true suffixes is impressive, it is usually important to limit false positives in privacy attacks because high false positive rates severely degrade trust, render the leakage signal unreliable, and can lead to overestimating privacy risk \cite{carlini2022membership}.

This is especially important in our attack, since the benchmark authors mention that the examples are designed to be somewhat easy-to-extract \cite{lmextractionchallenge2021}. This means that actual applications of targeted data extraction might return higher false-positive rates. 

Previous work explored limiting false extractions by defining a budget of false alarms, after which the attack stops accepting further predictions \cite{yu2023bag}. Although this approach is not directly applicable in adversarial settings where the attacker lacks access to ground truth suffixes, it highlights the utility cost associated with setting a false-extraction budget at a fixed level. In contrast, we systematically study the broader trade-offs and assess the effectiveness of different membership inference ranking methods under these conditions. Specifically, we apply a final thresholding step based on MIA scores to determine whether to accept or reject top-ranked suffixes.

\subsection{Methodology}
This step follows the same procedure as in standard MIA benchmarks. For a given generation strategy, we take the top-1 ranked suffix ($s$) for each one of the 1000 prefixes ($p$), ranked using likelihood scores. These suffixes were initially predicted as verbatim extraction and contain incorrect generations as explained before. Each one of these $(p, s)$ pair is then scored using the different MIA metrics. This score is used to predict whether the pair is a true extraction (training data member) or a false positive (non-training data member). We use the same scoring approach as the one used in the suffix ranking stage, which focuses on the probabilities of the suffix tokens. In Appendix C we briefly explore an alternative scoring approach that incorporates the probabilities of the prefix tokens.

To evaluate the performance of this step, we use standard MIA evaluation metrics. The metrics are:

\noindent\textbf{AUROC (Area Under the ROC Curve):} Following standard evaluation protocols for membership inference [22], [41], we use AUROC to measure the attack's ability to distinguish true extractions from false positives. This metric is defined as the area under the ROC curve, which plots the True Positive Rate against the False Positive Rate at varying decision thresholds. This is equivalent to the probability that a randomly selected true extraction is assigned a higher membership score than a randomly chosen incorrect guess (false positive), with 50\% representing random guessing.
\par\medskip
\noindent\textbf{TPR@5\%FPR} (True Positive Rate at 5\% False Positive Rate): The proportion of true extractions correctly identified when the threshold is set such that 5\% of incorrect guesses are misclassified as true.
\par\medskip
\noindent\textbf{FPR@95\%TPR }(False Positive Rate when TPR is 95\%): The false positive rate when the threshold is set to correctly identify 95\% of true extractions.
\par\medskip

\subsection{Results}
% This is the NEW discussion for Table V.
The results for the extraction confirmation stage, presented in Table~\ref{tab:mia_confirmation_all_gens}, show that a dedicated classification step can effectively filter false positives. In this task, the \textsc{S-ReCaLL} method consistently demonstrates the highest performance, achieving AUROC scores between 87.9\% and 91.0\% across all generation strategies. 

However, unlike some standard MIA benchmarks, the improvement over other methods is limited rather than dramatic. For example, using the Composite generation strategy, \textsc{S-ReCaLL} achieves an AUROC of 87.9\% and a TPR of 40.9\% at 5\% FPR. While this is the top score, other methods perform competitively, with the \textsc{Likelihood} baseline achieving an AUROC of 82.6\% and \textsc{Min-K\%} reaching 83.9\%. This indicates that while some methods offer an advantage for confirming a high-likelihood candidate, the raw likelihood score remains a surprisingly robust baseline even in this binary classification setting. 

\subsection{Tuning Hyperparameters for MIAs}

\begin{figure*}[t]
    \centering
    \includegraphics[width=0.6\columnwidth]{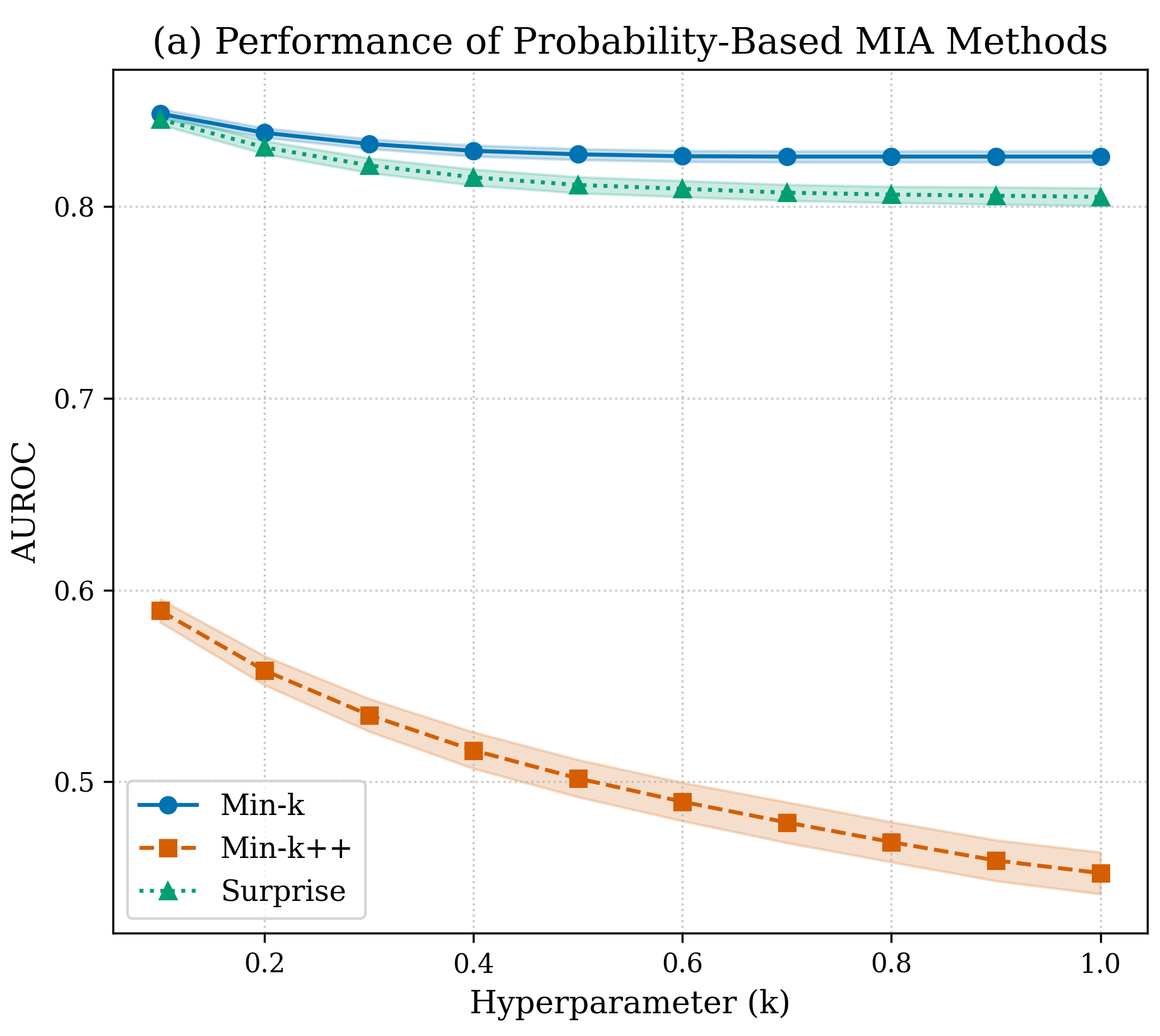}
       \hspace{0.3\columnwidth}
    \includegraphics[width=0.6\columnwidth]{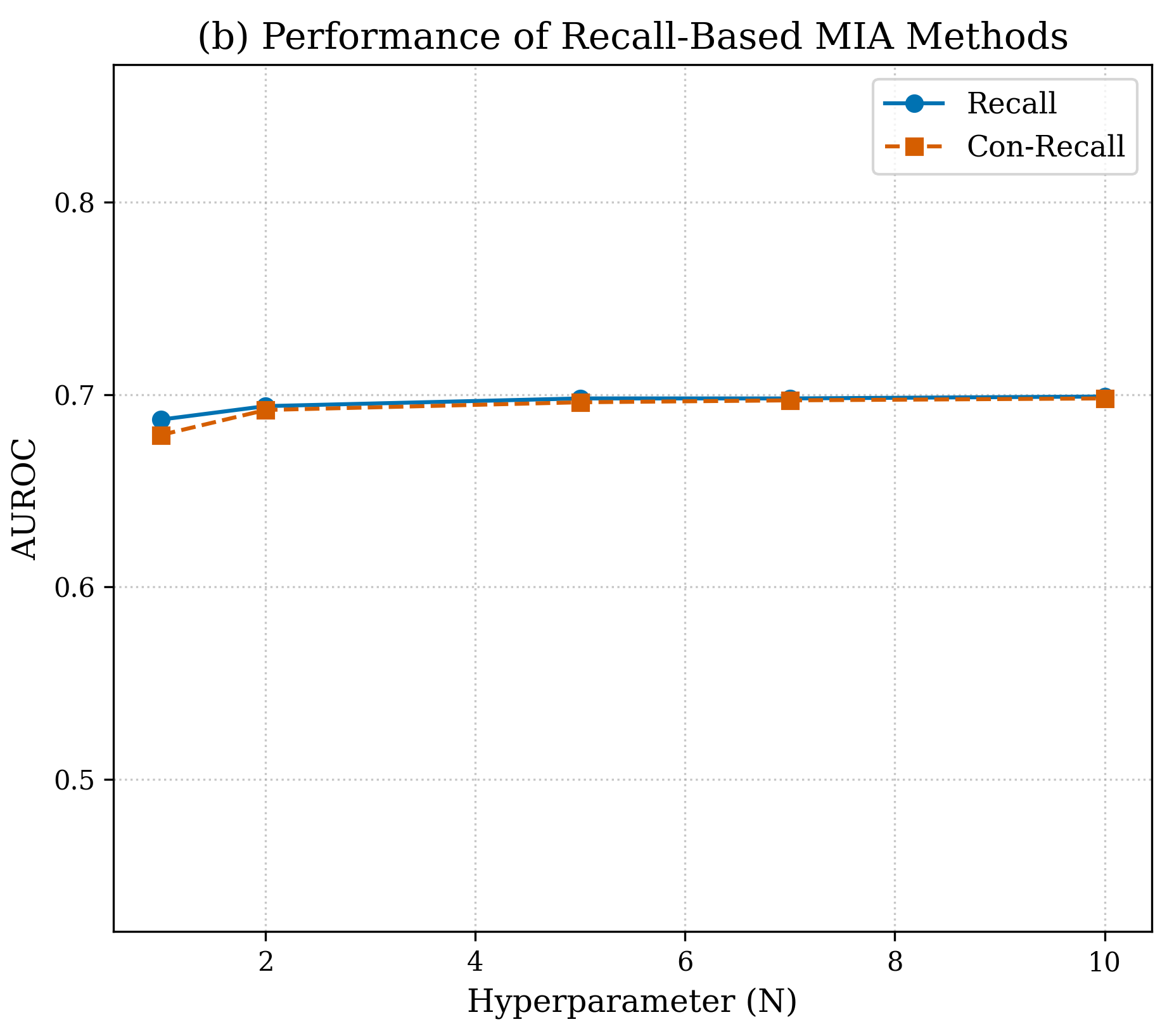}

    \caption{Performance of various MIA methods as their hyperparameters are tuned in a standard MIA setting.}
    \label{fig:hyperparameter_tuning}
\end{figure*}

Several of the evaluated MIA methods rely on hyperparameters that can significantly influence their performance. To illustrate this, we evaluated the impact of these parameters in a standard MIA setting, separate from the extraction pipeline. In particular, we examine how the performance of Min-K, Min-K++, and SURP changes with the threshold~$k$, and how ReCaLL and Con-ReCaLL are affected by the number of conditioning prefixes~$N$. Figure~\ref{fig:hyperparameter_tuning} shows the AUROC performance of these methods as their key hyperparameters are adjusted.

The results highlight that certain methods retain subpar performance even after tuning. In Figure~\ref{fig:hyperparameter_tuning}(a), the probability-based methods show distinct behaviors. For probability-based metrics, \textsc{Min-K++} consistently underperforms across thresholds and shows degradation with higher values of~$k$. Min-K and SURP perform strongly at lower values of~$k$. Both ReCaLL and Con-ReCaLL show only slight improvements when increasing the number of conditioning prefixes, and their performance remains below the top-performing methods.

\subsection{Results on the Extended Subset}
We also applied the MIA-based thresholding step to the top-1 outputs generated from the extended 15,000-pair dataset. To ensure the validity of our evaluation and rule out trivial distribution shifts, we additionally trained a model-less Bag-of-Words (BoW) classifier following the recommendations of Meeus et al. \cite{meeus2025sok}.

The results, shown in Table \ref{tab:extended_mia_confirmation}, are consistent with the findings from the 1,000-pair subset. On this larger dataset, S-ReCaLL again achieves the highest performance, with an AUROC of 87.5\% and a TPR@5\%FPR of 44.8\%. Notably, the Bag-of-Words baseline achieves an AUROC of only 64.2\%, significantly lower than the model-based methods.

Unlike the ranking stage where most methods performed almost identically, here we see a more distinct separation. While S-ReCaLL is the top performer, several other methods, including Min-K\% (AUROC 84.6\%) and Likelihood (AUROC 83.3\%), remain competitive. This indicates that raw model confidence remains a strong signal for this classification task. Methods like Lowercase, ReCaLL, and especially Min-K\%++ perform poorly, confirming their unsuitability for this confirmation step.

\begin{table}[b]
\centering
\caption{MIA Performance for Extraction Confirmation on the Extended Dataset.}
\label{tab:extended_mia_confirmation}
\begin{adjustbox}{max width=\columnwidth}
\begin{tabular}{l c c c}
\toprule
\textbf{MIA Method} & \textbf{AUROC (\%)} & \textbf{TPR@5\%FPR (\%)} & \textbf{FPR@95\%TPR (\%)} \\
\midrule
Bag of Words      & 64.2 & 16.3 & 86.9 \\
Likelihood        & 83.3 & 39.4 & 53.0 \\
Zlib              & 82.9 & 40.4 & 54.9 \\
Outlier           & 80.5 & 32.6 & 55.8 \\
High Confidence   & 83.4 & 40.0 & 53.0 \\
S-ReCaLL          & \textbf{87.5} & \textbf{44.8} & \textbf{43.0} \\
ReCaLL            & 69.6 & 17.0 & 79.5 \\
Lowercase         & 68.3 & 13.5 & 85.5 \\
Con-ReCaLL        & 68.6 & 16.9 & 82.6 \\
Min-K\% (k=0.2)   & 84.6 & 41.1 & 48.9 \\
Min-K\%++ (k=0.2) & 58.5 &  6.3 & 80.8 \\
SURP (k=0.4)      & 82.1 & 39.4 & 58.3 \\
\bottomrule
\end{tabular}
\end{adjustbox}
\end{table}

\subsection{Aggregating Membership Signals with Boosting}

To investigate whether the various MIA metrics capture distinct aspects of memorization, we explored an ensemble approach. We hypothesized that while individual metrics are likely correlated, they might not be perfectly aligned, potentially offering a slight advantage when combined.

To test this, we treated the confirmation step as a supervised regression task. We trained an AdaBoost learner using the scores from all computed MIA metrics as input features to predict a continuous membership confidence score. The dataset was split into an 80\% training set and a 20\% test set. To derive classification metrics, we applied a standard threshold ($0.5$) to the output, effectively classifying the sample based on the predicted confidence.

Table \ref{tab:adaboost_results} compares the performance of this ensemble approach against the baseline and the best-performing individual metric.

The results demonstrate that aggregating these metrics yields a consistent, albeit modest, performance boost. With the ensemble, AUROC improves to 0.913 and classification accuracy rises to 80.0\%, representing a 1.6 percentage point increase over the best individual method. This improvement suggests that while the various MIA metrics likely target similar probability-based signals, they are not perfectly correlated. The ensemble is able to leverage these slight variances to better distinguish correct extractions, though the substantial overlap in the underlying signal limits the extent of the gain.

However, it is important to note that the practical utility of this ensemble approach is limited as the training  requires a labeled dataset of valid and invalid extractions from the target distribution. In a realistic targeted extraction attack, an adversary typically lacks access to sufficient ground-truth samples to train a supervised model. 

\begin{table}[t]
\centering
\caption{Comparison of Individual MIA Metrics vs. Ensemble Regressor}
\label{tab:adaboost_results}
\begin{tabular}{l c c}
\toprule
\textbf{Method} & \textbf{Accuracy} & \textbf{AUROC} \\
\midrule
Likelihood (Baseline) & 0.720 & 0.823 \\
Suffix ReCaLL (Best Single) & 0.784 & 0.873 \\
AdaBoost Ensemble & \textbf{0.800} & \textbf{0.913} \\
\bottomrule
\end{tabular}
\end{table}
\section{Targeted Extraction from Fine-Tuned Models}
\label{sec:finetuning}

While the previous sections analyzed extraction from pre-trained base models, a fundamental limitation in evaluating privacy risks on such models is the lack of transparency regarding their training data. For most state-of-the-art LLMs, the training corpora are not open-source, making it difficult for researchers to establish the "ground truth" of memorization---specifically. By performing controlled fine-tuning experiments, we can precisely control the injection of sensitive data and evaluate the attack against a wider range of model architectures.

\subsection{Experimental Setup}
To simulate a realistic privacy leakage scenario, we utilize the Enron Email Dataset \cite{klimt2004enron}, a standard benchmark frequently employed in previous studies to evaluate privacy leakage from large language models \cite{nakka2024pii, cheng2025effective, lukas2023analyzing}. We constructed a training set designed to measure the impact of data repetition on memorization. The dataset consists of a background corpus of 14,500 general email samples. Embedded within this corpus are 2,000 target samples, each consisting of an email body containing a phone number and the preceding contents of the email.

To evaluate 
k-eidetic memorization during fine-tuning, we varied the number of times each target sample was repeated in the training set. Specifically:

\begin{itemize}
\item 1 repetition: 1,000 unique samples.
\item 2--5 repetitions: 250 unique samples for each repetition level (2, 3, 4, and 5).
\end{itemize}

We fine-tuned two open-weight models: Llama-3.2-1B \cite{dubey2024llama} and Qwen-2.5-1.5B \cite{qwen25}. The models were trained using Low-Rank Adaptation (LoRA) \cite{hu2021lora} to mimic parameter-efficient fine-tuning workflows common in industry. The extraction attack was performed by prompting the fine-tuned models with the email prefix (text preceding the phone number) and using greedy decoding to generate the continuation. A successful extraction is defined as the exact reproduction of the ground-truth phone number digits.

\subsection{Impact of Repetition on Memorization}
We first evaluate the extraction success rate, which is the percentage of phone numbers correctly recovered by the model given the prefix. As shown in Table \ref{tab:finetune_extraction}, even a single occurrence in the fine-tuning data poses a significant privacy risk.

The attack on Llama-3.2-1B successfully extracts 33.5\% of phone numbers that appeared only once, while the attack on Qwen-2.5-1.5B extracts 44.8\%. As repetitions increase, the leakage rate grows significantly, with Qwen recovering over 94\% of samples that appeared five times. This confirms that fine-tuning on sensitive data, even with minimal repetition, renders LLMs highly susceptible to targeted extraction attacks, and that as expected, training data repetition increases the privacy risk.

\begin{table}[t]
\centering
\caption{Extraction Success Rate by Repetition Count}
\label{tab:finetune_extraction}
\begin{tabular}{c c c}
\toprule
\textbf{Repetitions} & \textbf{Llama-3.2-1B (\%)} & \textbf{Qwen-2.5-1.5B (\%)} \\
\midrule
1   & 33.5 & 44.8 \\
2   & 37.6 & 60.8 \\
3   & 68.0 & 87.6 \\
4   & 65.2 & 90.0 \\
\textbf{5}   & \textbf{73.6} & \textbf{94.4} \\
\bottomrule
\end{tabular}
\end{table}

\subsection{Membership Inference for Fine-Tuning}
In an adversarial setting, the attacker generates a candidate phone number but does not know if it is correct. We applied the MIA techniques discussed in Section IV to distinguish between correct extractions (memorized data) and incorrect ones. To ensure the robustness of our results, we also included a model-agnostic Bag-of-Words (BoW) classifier as a baseline.

Table \ref{tab:finetune_mia} presents the AUROC scores for validating phone numbers generated via greedy decoding. First, we observe that the Bag-of-Words baseline achieves an AUROC of only 0.642, significantly lower than the model-based methods. This confirms that the high performance of MIAs is driven by the models' specific memorization patterns rather than simple distributional shifts in the text.

Comparing the model-based methods, the results are consistent with our previous findings on pre-trained models. The baseline performs remarkably well, achieving AUROC scores of 0.906 for Llama-3.2 and 0.913 for Qwen-2.5. More complex methods fail to achieve substantial or consistent gains over this baseline. For instance, while S-ReCaLL performs slightly better on Qwen (0.927), it actually underperforms on Llama (0.871). Similarly, Min-K\% does not surpass the baseline. This suggests that in the context of targeted extraction, raw model confidence remains a reliable signal for identifying memorized content.

\begin{table}[t]
\centering
\caption{AUROC of MIA Methods in Distinguishing Correct Extractions (Fine-Tuned Models)}
\label{tab:finetune_mia}
\begin{tabular}{l c c}
\toprule
\textbf{MIA Method} & \textbf{Llama-3.2-1B} & \textbf{Qwen-2.5-1.5B} \\
\midrule
Likelihood       & 0.906 & 0.913 \\
Zlib Entropy     & 0.905 & 0.912 \\
High Confidence  & \textbf{0.908} & 0.921 \\
Min-K\% (k=0.2)  & 0.892 & 0.873 \\
Min-K\%++ (k=0.2)& 0.868 & 0.749 \\
SURP (k=0.4)     & 0.894 & 0.917 \\
S-ReCaLL         & 0.871 &\textbf{0.927} \\
\bottomrule
\end{tabular}
\end{table}
\section{Discussion}
This paper is organized around two main parts of the data extraction pipeline. The first part provides a systematic benchmark of MIA techniques as ranking functions within the standard two-stage pipeline. The second part evaluates the use of MIA-based filtering to prune false positives from the final extracted output.

In the initial ranking stage, many advanced MIA methods, despite their conceptual sophistication and computational overhead, provide only marginal gains over the baseline likelihood score. As shown in Table \ref{tab:mia_ranking_all_gens}, the raw probability assigned by the model is a remarkably robust signal for distinguishing the correct suffix from a pool of diverse, machine-generated candidates.

A similar trend is observed in the MIA-based confirmation stage: most methods yield only marginal gains over the baseline likelihood, unlike the gains achieved in MIA benchmarks used in related work.

We also notice that both ReCall and Con-ReCall return subpar scores, despite achieving state-of-the-art performance on the WikiMIA benchmark \cite{wang2024con}. While S-ReCaLL  achieves AUROC scores approaching 90\% (Table \ref{tab:mia_confirmation_all_gens}). This might be due to S-ReCaLL leveraging prior knowledge unique to the targeted extraction framework, which is knowing that the 50-token prefix is part of the training data, allowing us to use the prefix itself as the conditioning data rather than using generic external prefixes. Since the candidate suffix is generated specifically to complete this prefix, the true prefix inherently provides a stronger conditioning context than any unrelated generic text. However, the observed score difference indicates that the substantial probability shift introduced by conditioning on any member or non-member data in MIA benchmarks (like WikiMIA) may not be universally observed across membership inference settings.

A nuanced theme emerges from our results. Standard benchmarks relying on post-hoc data collection (WikiMIA) show that MIA methods outperform the baseline by wide margins, but are criticized for the temporal shift in the data distribution. On the other hand, benchmarks that minimize distribution shift (MIMIR) show that MIA attacks perform near random guessing. Our results show that in the extraction pipeline, membership inference baseline performs really well, while other MIA methods fail to achieve substantial improvements. The extraction pipeline isolates verbatim memorization by requiring the attack to distinguish the true suffix from high-likelihood, model-generated plausible alternatives. This setup effectively neutralizes distributional artifacts, forcing the MIA to detect genuine memorization signals. Consequently, we observe that MIA results vary significantly according to the data domain and attack settings, demonstrating that current benchmarks do not reliably generalize to practical extraction scenarios. While benchmarks that evaluate MIAs under the hardest settings indicate that it is difficult to design an attack that consistently generalizes across different member--non-member distributions, our results show that these benchmarks don't account for privacy leaks that could occur in specific attack settings.  

\section{Conclusion}
This work presented a systematic study of how Membership Inference Attacks (MIAs) integrate into targeted data extraction from large language models. By evaluating a broad set of ranking and confirmation methods across different generation strategies, dataset scales, and model sizes, we provided a comprehensive assessment of their effectiveness.
Our results highlight two key insights. First, complex MIA techniques yield only marginal improvements over simple likelihood-based ranking in the candidate selection stage, suggesting that the added complexity may not translate into substantial benefits. Second, MIAs become more useful in the confirmation stage, where methods such as S-ReCaLL can reduce false positives and improve the reliability of extraction outcomes.
More broadly, this study reinforces recent findings that MIAs are neither universally strong nor uniformly weak, but instead exhibit highly inconsistent performance that depends on model scale, data characteristics, and evaluation setup. Future work should account for this context-dependency by systematically studying MIAs across diverse datasets, architectures. Future work can also explore tailoring attacks to specific setups rather than seeking membership inference that generalizes to all data domains and threat models.

\section*{LLM Usage Considerations}
LLMs were slightly used for editorial purposes in this manuscript, and all outputs were inspected and modified by the authors to ensure accuracy and originality
\section*{Ethical Statement}
This study relies exclusively on publicly available datasets (The Pile) and open-source models (GPT-Neo), without accessing or utilizing any private or sensitive user data. Its primary objective is to investigate, quantify, and benchmark techniques for training data extraction from LLMs, with the broader goal of assessing the privacy risks inherent in current models and guiding the design of more secure and privacy-preserving AI systems.

Although this study explores methods that enhance data extraction, its purpose is not to enable malicious use but to advance the scientific understanding of LLM vulnerabilities. By identifying effective extraction techniques, we aim to encourage the development of stronger defenses, privacy-preserving training methods, and more rigorous auditing practices for models trained on sensitive data. We emphasize the importance of responsible disclosure and the ethical study of such vulnerabilities as essential steps toward building safer and more trustworthy AI systems.

\bibliographystyle{IEEEtran}
\bibliography{arxiv/references_arxiv}
\clearpage
\onecolumn
\appendix

\subsection{Hyperparameter Configurations and Implementation Details}

This section provides the specific hyperparameter configurations and implementation details for the methods evaluated in this study. To improve clarity and consolidate information, the settings for both generation and ranking strategies are presented in a single table (Table \ref{tab:hyperparam_details}). The generation settings are based on the optimal configurations identified by Yu et al. \cite{yu2023bag} for the data extraction task. For ReCaLL methods, non-member data is sampled for pile\_cc split of MIMIR test set, and member data was sampled from a subset of 
the data extraction challenge disjoint from the data used in the experiments.

\begin{table}[!h]
\centering
\caption{Hyperparameter and Implementation Details}
\label{tab:hyperparam_details}
\begin{tabular}{@{}ll@{}}
\toprule
\textbf{Method / Strategy} & \textbf{Hyperparameter Setting and Implementation Details} \\ \midrule
\multicolumn{2}{l}{\textit{\textbf{Generation Strategies (based on \cite{yu2023bag})}}} \\
\midrule
Nucleus Sampling        & We use p = 0.6. \\
Temperature Sampling    & We use T = 0.3 \\
Typical p Sampling      & We use p = 0.6 \\
Top-k Sampling          & We use k = 10. \\
Repetition Penalty      & We use penalty = 1.1  \\
Composite               & We use a combination of the above: top-p=0.8, top-k=24, temperature=0.58, and rep-penalty=1.04, and typical-p=0.9. \\
\midrule
\multicolumn{2}{l}{\textit{\textbf{Ranking Methods (MIAs)}}} \\
\midrule
Min-K\%                 & The threshold for the lowest probability tokens is set to k = 0.2.  \\
Min-K\%++               & Following the Min-K\% setup, the threshold is also set to k = 0.2. \\
SURP                    & The surprising token threshold is set to k = 0.4  \\
High Confidence         & A token is considered high-confidence if its probability exceeds $\tau_{hc} = 0.9$, with a bonus of $b=1.0$. \\
ReCaLL                  & We use $N = 1$ non-member prefixes. \\
Con-ReCaLL              & We use $N = 1$ non-member prefixes and set the weighting factor $\gamma=1.0$.  \\ 
\midrule
\multicolumn{2}{l}{\textit{\textbf{Bag-of-Words (BoW) Baseline}}} \\
\midrule
Classifier              & Random Forest (500 trees). \\
Tree Constraints        & Max depth = 2, min samples per leaf = 10. \\
Features                & 1-grams appearing in $\geq$ 5\% of training documents. \\
Evaluation              & 80/20 train-test split, averaged over 5 runs. \\
\midrule
\multicolumn{2}{l}{\textit{\textbf{Fine-tuning (LoRA)}}} \\
\midrule
LoRA Configuration      & Rank $r=16$, Alpha $\alpha=32$, Dropout $p=0.05$. \\
Target Modules          & q\_proj, v\_proj, k\_proj, o\_proj, gate\_proj, up\_proj, down\_proj. \\
Training Parameters     & Learning Rate $2\text{e-}4$, Batch Size 4, Epochs 3. \\
Optimizer               & Paged AdamW (32-bit) with FP16 mixed precision. \\
Dataset & Sampled from enron emails, with each sample approximately 200 tokens long\\
\bottomrule
\end{tabular}
\end{table}

\subsection{Ablation Study: Scoring Suffix vs. Full Sequence}
In the main paper, the extraction confirmation stage (Section V) scores each top-1 candidate suffix to classify it as a true or false extraction. An alternative approach is to score the entire sequence, i.e., the concatenation of the original prefix $p$ and the candidate suffix $s$, denoted $(p,s)$. This ablation study compares these two approaches to determine which provides a stronger signal for membership. We hypothesized that scoring the full sequence could be more effective, as the model's behavior on the known-member prefix $p$ might provide additional context.

We repeated the extraction confirmation experiment using the GPT-Neo 1.3B model and candidates from the Composite generation strategy. Table \ref{tab:ablation_scoring_app} compares the performance of key MIA methods when scoring only the suffix $s$ versus the full sequence $(p,s)$.

\begin{table}[!h]
\centering
\caption{MIA Confirmation Performance: Suffix-Only vs. Full Sequence Scoring (\%)}
\label{tab:ablation_scoring_app}
\adjustbox{max width=\columnwidth}{
\begin{tabular}{l|ccc|ccc}
\toprule
& \multicolumn{3}{c|}{\textbf{Suffix-Only Scoring}} & \multicolumn{3}{c}{\textbf{Full Sequence Scoring}} \\
\cmidrule(lr){2-4} \cmidrule(lr){5-7}
\textbf{MIA Method} & AUROC & TPR@5 & FPR@95 & AUROC & TPR@5 & FPR@95 \\
\midrule
Likelihood      & 82.6 & 37.2 & 52.3 & 70.0 & 19.2 & 81.0 \\
Zlib            & 82.2 & 36.6 & 53.2 & 67.8 & 19.2 & 86.4 \\
Outlier          & 79.6 & 25.6 & 54.9 & 69.5 & 20.0 & 81.4 \\
High Confidence & 82.8 & 36.8 & 52.4 & 70.1 & 19.6 & 80.6 \\
Min-K\%         & 83.9 & 38.3 & 49.9 & 68.2 & 19.4 & 88.7 \\
Min-K\%++       & 55.7 & 6.5  & 84.1 & 62.6 & 7.2  & 88.2 \\
SURP            & 81.7 & 37.4 & 57.1 & \textbf{76.2} & \textbf{21.0} & \textbf{70.9} \\
\bottomrule
\end{tabular}
}
\end{table}

The results, presented in Table \ref{tab:ablation_scoring_app}, show a surprising trend. Contrary to our initial hypothesis, scoring the full sequence $(p,s)$ leads to a noticeable degradation in performance for most standard MIA methods compared to scoring only the suffix $s$. For instance, the AUROC for the Likelihood method drops from 82.6\% to 70.0\%, and for Min-K\%, it decreases from 83.9\% to 68.2\%. The SURP method is a notable exception, showing a significant improvement in AUROC from 81.7\% to 76.2\%, although its performance still lags behind the top suffix-only methods.

\subsection{Interaction Analysis: Generation vs. Ranking}
\label{app:interaction_analysis}

While previous sections focused on identifying the optimal combinations of generation and ranking strategies, it is equally important to understand the stability of these methods. To analyze these interaction effects, we calculated the mean precision ($\mu_{\Mp}$) and variance ($\sigma^2_{\Mp}$) for each ranking method across all generation strategies, and conversely, for each generation strategy across the effective ranking methods.

Table \ref{tab:interaction_stats} presents these statistics side-by-side. We observe a distinct pattern in the stability of ranking methods. Most effective membership inference metrics, such as Likelihood, Zlib, S-ReCaLL, and Min-K\%, exhibit very similar variance profiles, ranging approximately between 4.3 and 5.9. This consistency suggests that these scoring mechanisms behave similarly regardless of the specific sampling strategy used to generate the candidates. The only significant deviations occur with Lowercase and Min-K\%++, which show drastically higher variances of 57.91 and 27.33 respectively, confirming their instability and sensitivity to specific generation artifacts.

\begin{table}[!h]
\centering
\caption{Mean and Variance of Extraction Precision ($\Mp$)}
\label{tab:interaction_stats}
\footnotesize
% Increase column padding (default is usually ~6pt). 
% 10pt makes the table wider naturally without forcing a stretch.
\setlength{\tabcolsep}{10pt} 

\begin{minipage}[t]{0.48\linewidth} 
    \centering
    \textbf{Ranking Stability Across Generations}
    \vspace{0.1cm}
    
    \begin{tabular}{l c c}
    \toprule
    \textbf{Ranker} & \textbf{Mean} & \textbf{Var} \\
    \midrule
    Likelihood      & 48.97 & 5.40 \\
    Zlib            & 48.55 & 4.29 \\
    Outlier         & 46.90 & 8.05 \\
    SURP            & 47.25 & 4.53 \\
    High Conf.      & 48.73 & 4.67 \\
    ReCaLL          & 48.92 & 5.88 \\
    S-ReCaLL        & 48.98 & 5.76 \\
    Lowercase       & 37.85 & 57.91 \\
    Con-ReCaLL      & 48.35 & 6.54 \\
    Min-K\%         & 49.07 & 5.55 \\
    Min-K\%++       & 43.40 & 27.33 \\
    \bottomrule
    \end{tabular}
\end{minipage}
\hfill % Pushes the minipages apart
\begin{minipage}[t]{0.48\linewidth}
    \centering
    \textbf{Generation Consistency Across Rankers} 
    \vspace{0.1cm}
    
    \begin{tabular}{l c c}
    \toprule
    \textbf{Generator} & \textbf{Mean} & \textbf{Var} \\
    \midrule
    Nucleus (top-p)    & 49.98 & 0.53 \\
    Temperature        & 50.32 & 0.36 \\
    Typical            & 49.60 & 0.79 \\
    Top-k              & 45.88 & 0.84 \\
    Rep. Penalty       & 44.42 & 0.81 \\
    Composite          & 50.28 & 0.50 \\
    \bottomrule
    \end{tabular}
\end{minipage}
\end{table}
\subsection{Detailed Performance Metrics}
This section provides a more granular view of the results summarized in the main paper.

\subsubsection{Detailed Suffix Ranking Performance vs. Generation Count}
To provide a more granular view of the impact of the candidate pool size, we present a detailed breakdown of suffix ranking performance across different numbers of generated candidates (N). Table \ref{tab:detailed_ranking_vs_generations} shows the exact precision ($\Mp$\%) and Hamming distance ($\Mh$) for each MIA ranker. The results are generated using the Composite generation strategy. This detailed breakdown reinforces the observation that while more candidates improve overall performance, the relative advantage of advanced MIAs over the Likelihood baseline remains minimal across all pool sizes.

\begin{table*}[!h]
\centering
\caption{Detailed Suffix Ranking Performance Across Different Numbers of Generated Candidates (N).}
\label{tab:detailed_ranking_vs_generations}
\footnotesize
\setlength{\tabcolsep}{2pt}
\begin{tabular}{@{}l@{\hspace{0.5em}}||S[table-format=2.1]|S[table-format=2.2] @{\hspace{0.5em}} S[table-format=2.1]|S[table-format=2.2] @{\hspace{0.5em}} S[table-format=2.1]|S[table-format=2.2] @{\hspace{0.5em}} S[table-format=2.1]|S[table-format=2.2] @{\hspace{0.5em}} S[table-format=2.1]|S[table-format=2.2] @{\hspace{0.5em}} S[table-format=2.1]|S[table-format=2.2]@{}}
\toprule
\textbf{MIA Ranker} & \multicolumn{2}{c}{\textbf{N=1}} & \multicolumn{2}{c}{\textbf{N=5}} & \multicolumn{2}{c}{\textbf{N=10}} & \multicolumn{2}{c}{\textbf{N=20}} & \multicolumn{2}{c}{\textbf{N=50}} & \multicolumn{2}{c}{\textbf{N=100}} \\
\cmidrule(lr){2-3} \cmidrule(lr){4-5} \cmidrule(lr){6-7} \cmidrule(lr){8-9} \cmidrule(lr){10-11} \cmidrule(lr){12-13}
& {$\Mp$(\%)} & {$\Mh$} & {$\Mp$(\%)} & {$\Mh$} & {$\Mp$(\%)} & {$\Mh$} & {$\Mp$(\%)} & {$\Mh$} & {$\Mp$(\%)} & {$\Mh$} & {$\Mp$(\%)} & {$\Mh$} \\
\midrule
Likelihood        & 45.0 & 17.77 & 49.7 & 16.15 & 50.1 & 16.01 & 50.8 & 15.75 & 51.1 & 15.68 & 51.1 & 15.61 \\
Zlib              & 45.0 & 17.77 & 49.4 & 16.24 & 49.6 & 16.23 & 50.0 & 16.08 & 50.3 & 15.95 & 50.2 & 15.96 \\
Outlier            & 45.0 & 17.77 & 48.2 & 16.42 & 48.7 & 16.30 & 49.1 & 16.19 & 48.8 & 16.23 & 48.7 & 16.18 \\
High Confidence   & 45.0 & 17.77 & 49.8 & 16.13 & 50.5 & 15.90 & 50.6 & 15.87 & 50.4 & 15.87 & 50.7 & 15.74 \\
S-ReCaLL          & 45.0 & 17.77 & \textbf{50.0} & \textbf{15.98} & \textbf{50.5} & \textbf{15.75} & \textbf{51.2} & \textbf{15.46} & \textbf{51.0} & \textbf{15.41} & 51.1 & \textbf{15.40} \\
ReCaLL            & 45.0 & 17.77 & 49.6 & 16.10 & 50.3 & 15.86 & 51.0 & 15.58 & 50.8 & 15.59 & 50.7 & 15.62 \\
Lowercase         & 45.0 & 17.77 & 45.5 & 17.43 & 44.6 & 17.72 & 44.0 & 17.97 & 42.8 & 18.28 & 42.3 & 18.48 \\
CON-ReCaLL        & 45.0 & 17.77 & 49.2 & 16.17 & 50.0 & 15.92 & 50.5 & 15.76 & 50.3 & 15.85 & 50.3 & 15.88 \\
Min-K\%           & 45.0 & 17.77 & 49.9 & 16.08 & 50.4 & 15.93 & 51.1 & 15.71 & \textbf{51.4} & 15.61 & \textbf{51.5} & 15.47 \\
Min-K\%++         & 45.0 & 17.77 & 47.6 & 16.76 & 47.4 & 16.97 & 47.4 & 17.06 & 47.2 & 17.07 & 47.1 & 17.09 \\
SURP              & 45.0 & 17.77 & 48.7 & 16.43 & 49.0 & 16.45 & 49.0 & 16.60 & 48.6 & 16.65 & 48.6 & 16.66 \\
\bottomrule
\end{tabular}
\end{table*}
\subsubsection{Detailed Suffix Ranking Performance vs. Model Scale}
To supplement the summary results in Table \ref{tab:target_model_effect}, this section provides a comprehensive breakdown of ranking performance across all evaluated MIA methods for different model sizes. Table \ref{tab:detailed_ranking_vs_model_scale} details the precision ($\Mp$\%) and Hamming distance ($\Mh$) for the GPT-Neo 125M, 1.3B, 2.7B, and GPT-J 6B models. All results were generated using the Composite generation strategy with N=20 candidate suffixes per prefix. The data reinforces the conclusion that while larger models are more susceptible to extraction, the relative performance gains from using advanced MIA rankers over the Likelihood baseline remain consistently marginal.

\begin{table*}[t]
\centering
\caption{Detailed Suffix Ranking Performance Across Different Target Model Sizes}
\label{tab:detailed_ranking_vs_model_scale}
\footnotesize
\setlength{\tabcolsep}{2pt}
\begin{tabular}{@{}l@{\hspace{0.5em}}||S[table-format=2.1]|S[table-format=2.2] @{\hspace{0.5em}} S[table-format=2.1]|S[table-format=2.2] @{\hspace{0.5em}} S[table-format=2.1]|S[table-format=2.2] @{\hspace{0.5em}} S[table-format=2.1]|S[table-format=2.2]@{}}
\toprule
\textbf{MIA Ranker} & \multicolumn{2}{c}{\textbf{GPT-Neo 125M}} & \multicolumn{2}{c}{\textbf{GPT-Neo 1.3B}} & \multicolumn{2}{c}{\textbf{GPT-Neo 2.7B}} & \multicolumn{2}{c}{\textbf{GPT-J 6B}} \\
\cmidrule(lr){2-3} \cmidrule(lr){4-5} \cmidrule(lr){6-7} \cmidrule(lr){8-9}
& {$\Mp$(\%)} & {$\Mh$} & {$\Mp$(\%)} & {$\Mh$} & {$\Mp$(\%)} & {$\Mh$} & {$\Mp$(\%)} & {$\Mh$} \\
\midrule
Likelihood        & 19.8 & 30.56 & 50.8 & 15.75 & 58.7 & 12.19 & 70.6 & 7.71 \\
Zlib              & 19.8 & 30.62 & 50.0 & 16.08 & 58.3 & 12.28 & 70.2 & 7.82 \\
Outlier            & 19.5 & 30.66 & 49.1 & 16.19 & 57.5 & 12.49 & 70.1 & 7.90 \\
High Confidence   & \textbf{20.2} & 30.52 & 50.6 & 15.87 & 58.4 & 12.25 & \textbf{70.8} & 7.66 \\
S-ReCaLL          & 20.1 & \textbf{30.32} & \textbf{51.2} & \textbf{15.46} & 58.1 & 12.29 & 70.4 & 7.66 \\
ReCaLL            & 19.5 & 30.41 & 51.0 & 15.58 & 57.6 & 12.28 & 70.2 & 7.76 \\
Lowercase         & 16.0 & 32.41 & 44.0 & 17.97 & 53.3 & 14.06 & 63.3 & 9.97 \\
CON-ReCaLL        & 19.0 & 30.85 & 50.5 & 15.76 & 57.9 & 12.28 & 69.5 & 8.06 \\
Min-K\%           & 19.7 & 30.51 & 51.1 & 15.71 & \textbf{58.8} & \textbf{12.05} & 70.5 & \textbf{7.70} \\
Min-K\%++         & 18.1 & 31.49 & 47.4 & 17.06 & 56.5 & 12.97 & 65.8 & 8.95 \\
SURP              & 19.4 & 30.82 & 49.0 & 16.60 & 57.9 & 12.34 & 69.5 & 8.15 \\
\bottomrule
\end{tabular}
\end{table*}
\subsubsection{Detailed Suffix Ranking Performance vs. Pythia Model Scale}
To complement the analysis of the GPT-Neo family, we provide the detailed breakdown of ranking performance for the Pythia model suite. Table \ref{tab:detailed_ranking_pythia} presents the extraction precision ($\Mp$) and Hamming distance ($\Mh$) for Pythia-410M, Pythia-1.4B, Pythia-2.8B, and Pythia-6.9B. Consistent with the findings for GPT-Neo, larger models exhibit significantly higher extraction rates, yet the relative ranking of MIA methods remains stable, with methods like S-ReCaLL providing consistent but small improvements over the baseline.

\begin{table*}[!h]
\centering
\caption{Detailed Suffix Ranking Performance Across Different Pythia Model Sizes}
\label{tab:detailed_ranking_pythia}
\footnotesize
\setlength{\tabcolsep}{2pt}
\begin{tabular}{@{}l@{\hspace{0.5em}}||S[table-format=2.1]|S[table-format=2.2] @{\hspace{0.5em}} S[table-format=2.1]|S[table-format=2.2] @{\hspace{0.5em}} S[table-format=2.1]|S[table-format=2.2] @{\hspace{0.5em}} S[table-format=2.1]|S[table-format=2.2]@{}}
\toprule
\textbf{MIA Ranker} & \multicolumn{2}{c}{\textbf{Pythia 410M}} & \multicolumn{2}{c}{\textbf{Pythia 1.4B}} & \multicolumn{2}{c}{\textbf{Pythia 2.8B}} & \multicolumn{2}{c}{\textbf{Pythia 6.9B}} \\
\cmidrule(lr){2-3} \cmidrule(lr){4-5} \cmidrule(lr){6-7} \cmidrule(lr){8-9}
& {$\Mp$(\%)} & {$\Mh$} & {$\Mp$(\%)} & {$\Mh$} & {$\Mp$(\%)} & {$\Mh$} & {$\Mp$(\%)} & {$\Mh$} \\
\midrule
Likelihood        & 29.8 & 23.17 & 48.9 & 13.46 & 56.0 & 10.36 & \textbf{62.5} & 7.73 \\
Zlib              & 29.7 & 23.25 & 48.6 & 13.64 & 55.7 & 10.42 & 62.1 & 7.83 \\
Outlier           & 28.8 & 23.51 & 47.6 & 13.78 & 55.3 & 10.46 & 62.3 & 8.02 \\
High Confidence   & 29.6 & 23.23 & 49.0 & 13.43 & 55.9 & 10.37 & \textbf{62.5} & 7.73 \\
S-ReCaLL          & \textbf{30.2} & 23.22 & \textbf{49.2} & \textbf{13.24} & 55.7 & 10.29 & \textbf{62.5} & 7.80 \\
ReCaLL            & 29.9 & 23.21 & 49.1 & 13.31 & 55.7 & 10.31 & 62.3 & \textbf{7.69} \\
Lowercase         & 24.9 & 25.84 & 43.6 & 16.22 & 51.0 & 12.79 & 57.4 & 9.86 \\
CON-ReCaLL        & 23.3 & 26.45 & 41.1 & 17.69 & 49.9 & 13.18 & 56.9 & 9.77 \\
Min-K\%           & 29.9 & \textbf{23.10} & 49.0 & 13.41 & \textbf{56.3} & \textbf{10.21} & \textbf{62.5} & 7.76 \\
Min-K\%++         & 28.3 & 24.24 & 45.9 & 14.59 & 53.0 & 11.02 & 60.2 & 8.56 \\
SURP              & 29.2 & 23.65 & 47.9 & 14.07 & 54.8 & 10.79 & 62.0 & 7.93 \\
\bottomrule
\end{tabular}
\end{table*}

\end{document}